\documentclass{article}

     \PassOptionsToPackage{numbers, compress}{natbib}

     \usepackage[preprint]{neurips_2019}

\usepackage[utf8]{inputenc} 
\usepackage[T1]{fontenc}    
\usepackage{hyperref}       
\usepackage{url}            
\usepackage{booktabs}       
\usepackage{amsfonts}       
\usepackage{nicefrac}       
\usepackage{microtype}      

\usepackage{amsmath}
\usepackage{algorithm,algpseudocode}
\algnewcommand{\Inputs}[1]{%
  \State \textbf{Inputs:}
  \Statex \hspace*{\algorithmicindent}\parbox[t]{1\linewidth}{\raggedright #1}
}
\algnewcommand{\Output}[1]{%
  \State \textbf{Output:}
  \Statex \hspace*{\algorithmicindent}\parbox[t]{1\linewidth}{\raggedright #1}
}
\algnewcommand{\Cache}[1]{%
  \State \textbf{Cache:}
  \Statex \hspace*{\algorithmicindent}\parbox[t]{1\linewidth}{\raggedright #1}
}
\algnewcommand{\Notation}[1]{%
  \State \textbf{Notation:}
  \Statex \hspace*{\algorithmicindent}\parbox[t]{1\linewidth}{\raggedright #1}
}
\algnewcommand{\Initialize}[1]{%
  \State \textbf{Initialize:}
  \Statex \hspace*{\algorithmicindent}\parbox[t]{1\linewidth}{\raggedright #1}
}
\renewcommand{\Comment}[2][.5\linewidth]{%
  \leavevmode\hfill\makebox[#1][l]{//~#2}
}
\usepackage{graphicx}
\usepackage{caption}
\usepackage{hyperref}
\hypersetup{
    colorlinks=true,
    linkcolor=blue,
    filecolor=magenta,      
    urlcolor=cyan,
}

\usepackage{multirow}

\usepackage[position=bottom]{subfig}
\def\badscore#1{{\color{red}#1}}
\usepackage{arydshln}
\usepackage[export]{adjustbox}
\title{Variations on the Chebyshev-Lagrange\\Activation Function}

\author{
  Yuchen Li$^{1,2}$, Frank Rudzicz$^{1,2,3,4,5}$, Jekaterina Novikova$^1$\\
  $^1$: Winterlight Labs, $^2$: University of Toronto, $^3$: Surgical Safety Technologies \\
  $^4$: Li Ka Shing Knowledge Institute, St Michael's Hospital, $^5$: Vector Institute for Artificial Intelligence \\
  Toronto, ON, Canada \\
  ychnlgy.li@utoronto.ca, frank@cs.toronto.edu, jekaterina@winterlightlabs.com
}

\begin{document}
\maketitle
\begin{abstract}
  We seek to improve the data efficiency of neural networks and present novel implementations of parameterized piece-wise polynomial activation functions.
  The parameters are the $y$-coordinates of $n+1$ Chebyshev nodes per hidden unit and Lagrangian interpolation between the nodes produces the polynomial on $[-1, 1]$. We show results for different methods of handling inputs outside $[-1, 1]$ on synthetic datasets, finding significant improvements in capacity of expression and accuracy of interpolation in models that compute some form of linear extrapolation from either ends. We demonstrate competitive or state-of-the-art performance on the classification of images (MNIST and CIFAR-10) and minimally-correlated vectors (DementiaBank) when we replace ReLU or $\tanh$ with linearly extrapolated Chebyshev-Lagrange activations in deep residual architectures.
\end{abstract}
\section{Introduction}
\begin{figure}[h]
    \begin{minipage}{0.55\textwidth}
        Polynomial interpolation of Chebyshev nodes is one of the most data-efficient approaches for modelling a noise-free, $n$-differentiable function on the interval $[-1, 1]$\cite{interpolationintro}. The error between the approximating polynomial $P_{n-1}$ of degree at most $n-1$ and the true function $f$ is described by:
        
        \begin{equation}
            \label{chebyerror}
            |f(x)-P_{n-1}(x)| \leq \frac{1}{2^{n-1}n!} \max_{\xi\in[-1,1]}|f^{(n)}(\xi)| 
        \end{equation}
        
        The primary difficulty of applying this to neural networks is the treatment of inputs that may be noisy or outside of $[-1, 1]$. We solve this by learning the support of polynomials that map inputs to outputs. For every input unit $x_i$, we learn the positions of $n+1$ points such that the polynomial $P_n$ that forms from these points has output $P_n(x_i)$ that minimizes the final loss function. We fix the $x$-positions of the support points to be the Chebyshev nodes and learn the $y$-positions. Figure \ref{cifar-visualization} shows an example of this mechanism. 
    \end{minipage}\hspace{.04\textwidth}
    \begin{minipage}{0.48\textwidth}
        \centering
        \includegraphics[scale=0.42]{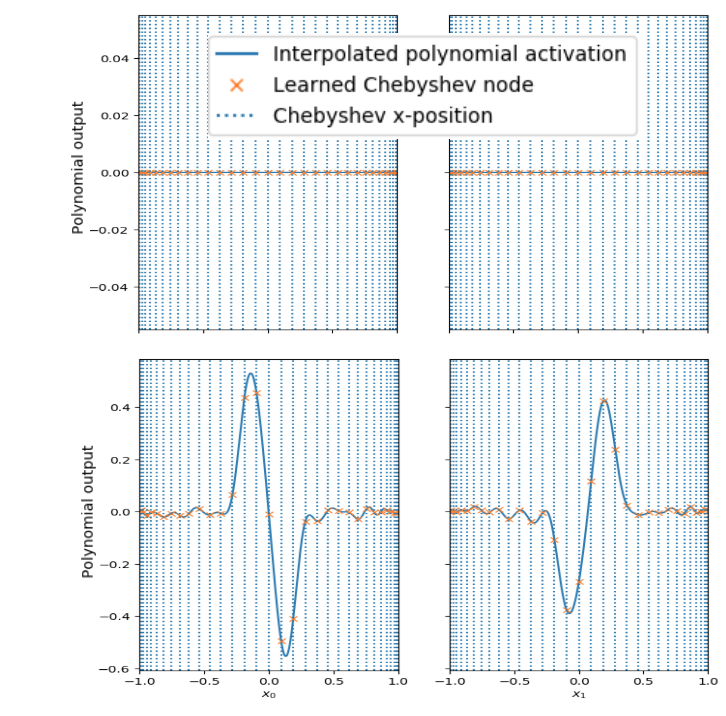}
        \caption{
             Chebyshev-Lagrange activations before \textit{(top row)} and after \textit{(bottom row)} recieving backpropagation for 100 epochs of training on CIFAR-10. We show the activations for the first \textit{(left column)} and second \textit{(right column)} elements of the last linear layer of a modified ResNet-32.
        }
        \label{cifar-visualization}
    \end{minipage}
\end{figure}
This idea has been explored with piece-wise polynomial activations where the model learns the weights for the Lagrangian basis functions of Chebyshev-Lobatto nodes \cite{piecewisepolynomial}, which is equivalent to learning the $y$-positions of the nodes. Apart from applying further piece-wise polynomials, there was no additional mechanism for handling inputs outside of $[-1,1]$. Therefore, the main novelty of our work is the linear extension of the polynomial from either end\footnote{Although we can linearly scale the Chebyshev nodes into an arbitrary range of $[a, b]$, it is generally unknown what range will definitely cover all inputs for any task. We also demonstrate in the results that if the input is not restrained within this stable region, the model quickly experiences exploding gradients. Therefore, we simply provide solutions for handling inputs outside of $[-1, 1]$.} using regression or extrapolation, which we find to be key in improving  results and attaining competitive performance with modern architectures on a variety of tasks.

We rationalize the basic Chebyshev-Lagrange method and our variations in Sections \ref{background} and \ref{implementation},  then describe the selection process for the best variation using synthetic datasets and demonstrate its practical value on real datasets in Sections \ref{datasets-and-arch} and \ref{results}, and finally offer explanations for its versatility and performance and discuss future work in Section \ref{discussion}.

\subsection{Background}\label{background}
Several existing architectures allude to the idea that parameterized piece-wise polynomial activations can improve the data efficiency of ReLU-based neural networks \cite{prelu,maxout,polynomialnn}. PReLU is a modified ReLU where the output slope of negative inputs is learned. It is claimed to be the reason for surpassing human performance on the task of classifying of ImageNet 2012 \cite{prelu}. Maxout is a parameterized activation function where the output is the maximum of a linear transformation \cite{maxout}. It performed better on various image classification tasks than similar architectures that used ReLU \cite{maxout}. In particular, maxout differed from ReLU in that it could learn to approximate simple piece-wise polynomials per hidden unit \cite{maxout}. We build upon these observations and explicitly implement polynomials with degrees greater than one, with the goal of complementing and competing with existing neural architectures in terms of data efficiency on a variety of tasks.

Weighted Chebyshev polynomials (WCP) and Lagrangian interpolation of Chebyshev nodes (CL) are two related ways of constructing polynomials with good interpolative properties within a chosen interval. We refer the reader to Levy \cite{interpolationintro} for introductory material on these methods. Original work in artificial neurons that used WCP demonstrated improved convergence speed \cite{chebyuniversal}, simplification of neural network operations \cite{chebysigmoid} and successful application in spectral graph convolutions \cite{graphcnn, structcnn, multigraphnet}. \cite{graphcnn} implemented spectral graph convolutions using a filter operation defined by the sum of weighted Chebyshev polynomials at the normalized graph Laplacian matrix $\Tilde{L}$ of all training data: $f(x) := \sum^n_{k=0} \theta_kT_k(\Tilde{L})x$, where $\theta_k$ are the parameters and $T_k$ are the Chebyshev basis polynomials defined by the recurrence relation $T_0(x) = 1,\; T_1(x) = x,\; T_i(x) = 2xT_{i-1}(x)-T_{i-2}(x)$ for $i\geq2$. Chebyshev polynomial modelling is an appropriate method here because the values of a normalized Laplacian matrix are contained in the interval $[-1, 1]$ and are constant per graph configuration. On the other hand, little work has been done on incorporating CL into neural networks. The main advantage of CL over WCP is that the former can be expressed in closed form and thus benefits from parallelized tensor operations. We will also explore variations of CL that take advantage of the straightforward gradients and nodes of CL for linear extrapolation. 

\section{Methods}
\label{methods}
\subsection{Implementation of the Chebyshev-Lagrange activation functions}\label{implementation}
We examine several variants of CL. $\tanh$ and prototype cosine-similarity \cite{deepconsensus} are two ways of compressing input spaces onto the interval $[-1, 1]$. Whereas the former computes this element-wise, the latter returns a vector representing the cosine-similarity of the entire input vector with each of its prototype parameters \cite{deepconsensus}. We refer to these as `restrictive' of the inputs to the polynomial. We also experiment with non-restrictive methods that linearly extrapolate (\textit{CL-extrapolate}) or regress (\textit{CL-regression}) the polynomial outside of $[-1, 1]$. They are expressed as:
\begin{equation}
    \label{chebyshevlagrange-formula}
    \sigma(v_i) =
    \begin{cases}
        m_{(-1)}v_i+(y_{n+1}-m_{(-1)}x_{n+1}) & v_i < -1 \\
        P_n(v_i) & -1 \leq v_i \leq 1 \\
        m_{(+1)}v_i + (y_1-m_{(+1)}x_1) & v_i > 1, \\
    \end{cases}
\end{equation}
where $v_i$ is the input unit, $m_{(-1)}$ and $m_{(+1)}$ are the slopes of the linear pieces and $P_n$ is a polynomial of degree $n$ defined by the Chebyshev nodes $x_1, x_2, \ldots, x_{n+1}$ and parameters $y_1, y_2, \ldots, y_{n+1}$. We choose to join the linear portions to the central polynomial. Table \ref{slope-difference} outlines the difference in linear slopes for extrapolation and regression. Table \ref{prototype-variants} summarizes all variants. We describe the parameters and implementation of Chebyshev-Lagrange activations in appendices \ref{appendix-chebyparams} and \ref{appendix-lagrangecalc} because they are not the main novelty. Since the backpropagation signal to $y_i$ is not directly proportional to the output of the activation, we aim for simplicity and choose to initialize all $y$-positions to zero as Figure \ref{cifar-visualization} shows.
\begin{table}[h]
    \centering
    \caption{Linear slope implementations for non-restrictive, piece-wise variants of Chebyshev-Lagrange activations (CL) at inputs less than ($m_{(-1)})$) or greater than $1$ ($m_{(+1)})$. For \textit{CL-regression}, we must also choose the hyperparameter $k$ for the number of nodes from either ends of the polynomial to use in regression. The index ordering of regression is reversed because $x_1$ is closest to $+1$ and $x_{n+1}$ is closest to $-1$ for Chebyshev nodes. Appendix \ref{appendix-lagrangegrad} shows an example of how to precompute the gradients of the Lagrangian bases on tensors.
    }
    \label{slope-difference}
    \begin{tabular}{c|c|c}
        \hline
        Model & $m_{(-1)}$ & $m_{(+1)}$ \\
        \hline
         Extrapolation & $P_n'(-1)$ & $P_n'(+1)$ \\
         Regression & $\text{Cov}(x_{n-k+1...n}, y_{n-k+1...n})/\text{Var}(x_{n-k+1...n})$ & $\text{Cov}(x_{1...k}, y_{1...k})/\text{Var}(x_{1...k})$ \\
         \hline
    \end{tabular}
\end{table}
\begin{table}[h]
    \centering
    \caption{
        Variations of the basic Chebyshev-Lagrange (CL) activation. The output is treated differently if the inputs are outside $[-1, 1]$ for piece-wise variants. We choose $n=3$ as the maximum degree of all CL polynomials and $k=2$ for the number of regression nodes.
    }
    \label{prototype-variants}
    \begin{tabular}{ccc}
        \hline
        Variant & Polynomial input & Output \\
        \hline
        Weighted Chebyshev polynomials & $\mathbb{R}$ & Smooth \\
        CL & $\mathbb{R}$ & Smooth \\
        $\tanh$-CL & $[-1, 1]$ & Smooth \\
        Prototype cosine-similarity \cite{deepconsensus} CL & $[-1, 1]$ & Smooth \\
        \textit{CL-regression} &  $\mathbb{R}$ & Piece-wise continuous \\
        \textit{CL-extrapolate} &  $\mathbb{R}$ & Piece-wise continuous \\
        \hline
    \end{tabular}
\end{table}
\subsection{Datasets and architectures}\label{datasets-and-arch}
\paragraph{Synthetic datasets.} To select the best CL variants from Table \ref{prototype-variants}, we train and test on 7 artificially generated datasets with different nonlinear input-output relationships. To generate data, the columns of matrices with random numbers sampled uniformly on $[-1, 1]$ combine according to the recipes listed in Table \ref{synthetic-description} to produce the target outputs of each dataset, which then receives Gaussian noise of 0.01 or 0.04 standard deviations to simulate field observations. We produce 2000 data points for $N=1000$ training and testing sets. See Appendix \ref{appendix-samplesynthetic} for sample visualizations of these datasets.

\begin{table}[h]
    \centering
    \caption{Recipes of the target output for each synthetic dataset. Subscripts of $x$ indicate the column index of the uniform random matrix input, which has exactly the number of required columns.}
    \begin{tabular}{c|c|c}
        \hline
         Dataset & Input type & Output recipe \\
         \hline
         Pendulum & Smooth & $-x_1x_2\sin(2\pi x_0)$ \\
         Arrhenius & Smooth & $x_1e^{-x_2x_0/4}$\\
         Gravity & Smooth & $x_1x_2x_3/(0.2+x_0^2)$\\
         Sigmoid & Smooth & $2x_1/(1+e^{-10x_2(x_0-x_3+0.5)})+x_4-0.5$ \\
         PReLU & Continuous & if $x_0<0$ then $0.1x_0x_1$ else $x_0x_2$\\
         Jump & Discontinuous & if $x_0 < x_1-3/4$ then $4x_2x_0$ else $0.1x_3((4x_2x_0)-x_2/2)$\\
         Step & Discontinuous & for $t$ in $[-0.8, -0.4, 0, 0.4, 0.8]$, return $t$ if $x_0 < t$. If done, return $0.8$.\\
         \hline
    \end{tabular}
    \label{synthetic-description}
\end{table}

The architectures are 32-width residual networks that use the different activations and properties described in Table \ref{synthetic-architectures}.
Weights of all fully-connected and convolution layers in this paper are initialized using the He uniform distribution though it likely benefits ReLU \cite{prelu} more than CL.
We train these variants for 300 epochs on each dataset using L1-loss, stochastic gradient descent (SGD) with batch size 32, 0.01 learning rate, 0.99 momentum, $1\times10^{-6}$ weight decay, cosine annealing learning rate scheduling \cite{cosine-anneal} and report the test performance in root mean square error (RMSE). Note that we choose L1-loss instead of mean squared error because the latter required careful selection of learning rates for datasets involving exponentials (Arrhenius, Sigmoid), large gaps in discontinuities (Jump) or inverses (Gravity) to avoid exploding gradients. We repeat all experiments across 10 random seeds.

\begin{table}[h]
    \centering
    \caption{
        Chebyshev-Lagrange (CL) architecture summary for the synthetic experiments. The base architecture is a 32-width residual network using fully connected layers. We include weighted Chebyshev polynomials (WCP) as a baseline. All CL activations have maximal degree $n=3$.
        The rows are grouped by the type of activation: controls, non-restrictive inputs, restrictive inputs and non-restrictive inputs with piecewise outputs. ReLU and $\tanh$ are vanilla activations. Cubic computes the third power of the input. Prototype cosine-similarity \cite{deepconsensus} (PCS) and $\tanh$ are two different ways of compressing inputs onto $[-1, 1]$ before passing it to CL (PCS-CL and $\tanh$-CL). \textit{CL-extrapolate} and \textit{CL-regression} compute linear extrapolation and regression from the two ends of the polynomial for any inputs outside of $[-1, 1]$.
    }
    \label{synthetic-architectures}
    \begin{tabular}{ccccc}
        \hline
        Activation & Residual blocks & Layers/Block & Parameters \\
        \hline
        ReLU & 3 & 1 & 3300 \\
        ReLU (2x depth) & 6 & 1 & 6500 \\
        ReLU (2x layers) & 3 & 2 & 6500 \\
        $\tanh$ & 3 & 1 & 3300 \\
        Cubic & 3 & 1 & 3300 \\
        \hline
        CL & 3 & 1 & 3800 \\
        WCP & 3 & 1 & 3800 \\
        \hline
        PCS-CL & 3 & 1 & 6785 \\
        $\tanh$-CL & 3 & 1 & 3800 \\
        \hline
        \textit{CL-regression} & 3 & 1 & 3800 \\
        \textit{CL-extrapolate} & 3 & 1 & 3800 \\
        \hline
    \end{tabular}
\end{table}

\paragraph{DementiaBank.}\label{methods-dementiabank} DementiaBank is a collection of 551 audio recordings of 210 subjects in various stages of cognitive decline \cite{dementiabank}. We filter for subjects with dementia or no diagnosis and count 178 recordings with the dementia label and 229 with no diagnosis. Per recording, we extract the 480 minimally-correlated features specified in \cite{heterogeneousdata}. As demonstrated in \cite{fraser-linguistic-alzheimers} and \cite{multiview-emb}, feature selection is critical for good performance on this dataset. For feature selection, we use the FamousPeople dataset, which is a collection of 543 audio recordings of 32 subjects in various stages of cognitive decline \cite{heterogeneousdata}. We replicate \cite{heterogeneousdata} by computing and normalizing the same 480 features used in DementiaBank for the recordings. We follow \cite{heterogeneousdata} and use the Boruta algorithm \cite{boruta} on FamousPeople to select the 66 final features listed in Appendix \ref{appendix-dementiabank-borutafeatures}. These features are used for DementiaBank classification.

DementiaBank classification experiments investigate the effect of replacing ReLU or $\tanh$ with the best-performing CL variant from the synthetic experiments. The base network is a 2-block, 6-layer residual network \cite{resnet}, with widths of 32 or 64 to control for hyperparameter optimization. As per \cite{fraser-linguistic-alzheimers, heterogeneousdata, normativedata}, the training and testing sets are produced from random 10-fold cross validation such that each testing fold does not contain samples of subjects existing in the training set. The models train for 300 epochs with cross entropy loss, SGD with batch size 32, 0.01 learning rate, 0.9 momentum, $1\times10^{-4}$ weight decay, cosine annealing learning rate scheduling \cite{cosine-anneal} and we report the average validation accuracy, sensitivity, specificity and F1 score for the 10-fold cross validation. Since we repeat the experiments across 30 random seeds, this amounts to 300 random cross validation folds.

\paragraph{MNIST and CIFAR-10.}\label{methods-imageclass} MNIST contains 60,000 training and 10,000 testing samples of black and white handwritten digits in $28\times28$ pixel arrays \cite{mnist}. CIFAR-10 contains 50,000 training and 10,000 testing samples of RGB images in $3\times32\times32$ pixel arrays \cite{cifar10}. The typical task for both datasets is to map the pixel inputs to one of the ten labelled classes. For MNIST, we pad images to $32\times32$ pixels and augment with 2-pixel translations. For CIFAR-10, we apply the typical data augmentation techniques of random 4-pixel translations and horizontal flips \cite{resnet, shake-shake}. The same architecture described in \cite{shake-shake} is used for both datasets, but we choose 14 layers instead of 26 and progressing widths of 16, 32, 64 and 128 instead of the original progression of 16, 64, 128 and 256. We also train for 100 instead 300 epochs. These architectural and training changes are meant to reduce the computation time of repeating all experiments across 3 random seeds and we expect lower final scores as a result. Shake-shake regularization is disabled 
for MNIST since we obtain no discernible benefit with it. We replicate these experiments using ReLU and \textit{CL-extrapolate} as the last 8 convolutional activations.

\begin{figure}[h]
    \begin{minipage}[t]{0.55\textwidth}\vspace{0pt}
        \subsection{Architecture details}
        \label{appendix-architectures}
        \begin{itemize}
            \item For the synthetic experiments, we chose a 5-layer, 3-block residual network. Each fully-connected layer maps $d$ input units to $d$ output hidden units. $\sigma$ refers to one non-linear activation function from Table \ref{synthetic-architectures}. Curved arrows indicate addition. We choose $d=32$. The number of input features $i$ was the exact number of features required for each recipe in Table \ref{synthetic-description}.
            \item For DementiaBank classification, we used a 6-layer, 2-block residual network. Each fully-connected layer maps $d$ input units to $d$ hidden units. We chose $\sigma$ to be one of ReLU, $\tanh$ or \textit{CL-extrapolate}. Curved arrows indicate averaging of the shortcut with the residual output. We choose $d=32$ or $64$. The number of input features was $i=66$.
            \item For image classification, we replicated the architecture in \cite{shake-shake} with modifications listed in Section \ref{methods-imageclass}. We use batch normalization (BN) following every convolution in every residual block. Convolution layers are denoted by kernel size, channel width and stride, if any. Shake-shake (SS) regularization \cite{shake-shake} was only used in CIFAR-10 but not MNIST. In MNIST, we simply averaged the outputs of the two chains. We chose $\sigma$ to be one of ReLU or \textit{CL-extrapolate}. Final $8\times8$ global pooling was used as per \cite{shake-shake}, followed by a fully-connected layer mapping the $128$ hidden units to the $10$ class predictions of either dataset. The total number of trainable parameters was 1.4 million for both ReLU and \textit{CL-extrapolate} variants.
        \end{itemize}
    
    \end{minipage}\hspace{.04\textwidth}
    \begin{minipage}[t]{0.48\textwidth}\vspace{0pt}
        \centering
        \includegraphics[scale=0.45]{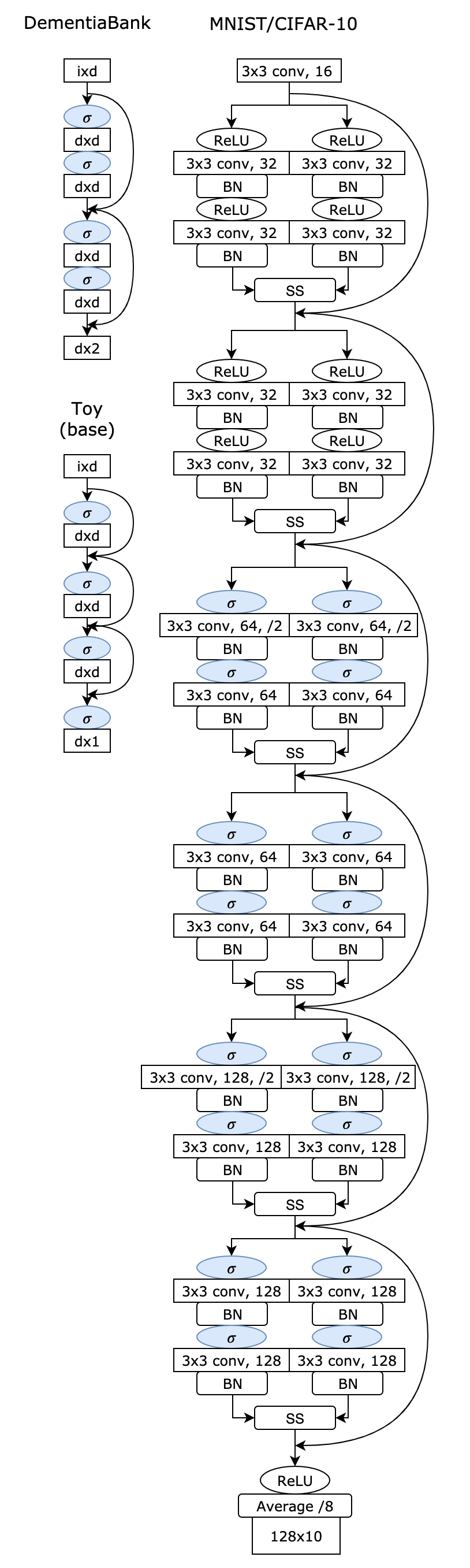}
        \label{architecture-details}
    \end{minipage}
\end{figure}

Appendix \ref{appendix-architectures} provides more details about the architectures.

\section{Results}
\label{results}
\subsection{Synthetic datasets}
Table \ref{synthetic-0.01} shows the RMSE mean and standard deviations for 10 repeated trials of each model in Table \ref{synthetic-architectures} on each dataset in Table \ref{synthetic-description} with 0.01 standard deviations of Gaussian noise. Given the same data points, models using Chebyshev-Lagrange variants that compute linear extrapolation (\textit{CL-extrapolate}) show similar or notable improvement of interpolation accuracy over all other classes of activation, even when the underlying functions are discontinuous (i.e., Jump and Step). Despite limiting the raw input in $[-1, 1]$, the non-restrictive Cubic, CL and WCP activations succumb to exploding gradients on the majority of these datasets.  

\begin{table}[h]
    \centering
    \caption{Root mean squared error (RMSE) for the smooth (\ref{synthetic-smooth-0.01}) and non-smooth (\ref{synthetic-nonsmooth-0.01}) synthetic datasets with 0.01 standard deviations of Gaussian noise. NaN indicates counts of gradient explosion. See Table \ref{synthetic-description} and \ref{synthetic-architectures} for descriptions of the datasets and models. Excluding rows containing any NaN, bold font marks the two most data-efficient methods while red font marks the four least data-efficient methods.}
    \label{synthetic-0.01}
    \subfloat[Smooth functions]{
        \label{synthetic-smooth-0.01}
        \begin{tabular}{c|c|c|c|c}
        \hline
        Model & Pendulum & Arrhenius & Gravity & Sigmoid \\
        \hline
        ReLU & $\badscore{0.1518\pm0.0297}$ & $\badscore{0.0054\pm0.0003}$ & $\badscore{0.098\pm0.069}$ & $\badscore{0.065\pm0.010}$ \\
        ReLU (2x depth) & $\badscore{0.1941\pm0.0207}$ & $\badscore{0.0052\pm0.0004}$ & $\badscore{0.187\pm0.115}$ & $\badscore{0.058\pm0.007}$ \\
        ReLU (2x layers) & $\badscore{0.1597\pm0.0203}$ & $\badscore{0.0048\pm0.0003}$ & $\badscore{0.062\pm0.049}$ & $\badscore{0.043\pm0.005}$\\
        $\tanh$ & $0.0327\pm0.0102$ & $0.0036\pm0.0005$ & $\badscore{0.075\pm0.059}$ & $\badscore{0.111\pm0.013}$\\
        Cubic & (10/10 NaN) & (4/10 NaN) & (10/10 NaN) & (10/10 NaN) \\
        \hline
        CL & (8/10 NaN) & $0.0029\pm0.0007$ & (10/10 NaN) & (10/10 NaN) \\
        WCP & (9/10 NaN) & $0.0032\pm0.0010$ & (10/10 NaN) & (10/10 NaN) \\
        \hline
        PCS-CL & $0.0348\pm0.0021$ & $0.0045\pm0.002$ & $\mathbf{0.022\pm0.002}$ & $0.035\pm0.010$ \\
        $\tanh$-CL & $\badscore{0.1642\pm0.0749}$ & $\badscore{0.0056\pm0.0015}$ & $0.048\pm0.009$ & $0.041\pm0.011$ \\
        \hline
        \textit{CL-regression} & $\mathbf{0.0208\pm0.0022}$ & $\mathbf{0.0035\pm0.0002}$ & $0.042\pm0.003$ & $\mathbf{0.027\pm0.004}$ \\
        \textit{CL-extrapolate} & $\mathbf{0.0113\pm0.0006}$ & $\mathbf{0.0030\pm0.0002}$ & $\mathbf{0.022\pm0.002}$ & $\mathbf{0.019\pm0.004}$ \\
        \hline
        \end{tabular}
    }

    \subfloat[Non-smooth functions]{
        \label{synthetic-nonsmooth-0.01}
        \begin{tabular}{c|c|c|c}
            \hline
            Model & Jump & PReLU & Step \\
            \hline
            ReLU & $\badscore{0.17\pm0.15}$ & $\badscore{0.0063\pm0.0006}$ & $\badscore{0.040\pm0.010}$ \\
            ReLU (2x depth) & $\badscore{0.17\pm0.12}$ & $\badscore{0.0079\pm0.0053}$ & $0.035\pm0.010$\\
            ReLU (2x layers) & $\badscore{0.12\pm0.03}$ & $0.0058\pm0.0004$ & $\mathbf{0.030\pm0.013}$\\
            $\tanh$ & $\mathbf{0.09\pm0.03}$ & $\badscore{0.0085\pm0.0041}$ & $\badscore{0.050\pm0.005}$\\
            Cubic & (10/10 NaN) & (9/10 NaN) & (7/10 NaN) \\
            \hline
            CL & (10/10 NaN) & $0.0071\pm0.0005$ & $0.109\pm0.018$\\
            WCP & (10/10 NaN) & $0.0066\pm0.0006$ & $0.091\pm0.025$\\
            \hline
            PCS-CL & $0.11\pm0.3$ & $\badscore{0.0060\pm0.0005}$ & $\badscore{0.069\pm0.007}$\\
            $\tanh$-CL & $\badscore{0.12\pm0.03}$ & $\badscore{0.0060\pm0.0008}$ & $\badscore{0.088\pm0.018}$\\
            \hline
            \textit{CL-regression} & $\mathbf{0.08\pm0.02}$ & $\mathbf{0.0045\pm0.0003}$ & $\mathbf{0.020\pm0.001}$ \\
            \textit{CL-extrapolate} & $\mathbf{0.09\pm0.02}$ & $\mathbf{0.0040\pm0.0003}$ & $\mathbf{0.030\pm0.002}$ \\
            \hline
        \end{tabular}
    }
\end{table}

In order to test if these particular CL variants do better than the other activations only in cases of small noise pertubation, we increase the Gaussian noise standard deviation to 0.04. Table \ref{synthetic-0.04-samples} in Appendix \ref{appendix-syntheticnoisy} visualizes the magnitude of perturbation this has on the inputs from $[-1, 1]$. Table \ref{synthetic-0.04} in Appendix \ref{appendix-syntheticnoisy} shows that \textit{CL-extrapolate} continues to perform best for smooth functions and achieves identical performance to the best vanilla activation variants on non-smooth functions, while \textit{CL-regression} degrades in performance with the increased noise. Non-restrictive activations continue to experience high frequencies of exploding gradients. Although PCS-CL achieved the lowest RMSE on $2/7$ datasets, it performed inconsistently on datasets with less noise. Therefore, we selected \textit{CL-extrapolate} to investigate in the following experiments on real datasets.

\subsection{Dementia detection}
Table \ref{chebyeffect-dementiabank} shows the consistent improvement in various metrics of binary classification from replacing ReLU or $\tanh$ with \textit{CL-extrapolate}. Table \ref{dementiabank-globalcomparison} indicates that these results are, to the best of our knowledge, the state-of-the-art for this task and dataset. Note that results from \cite{dementiabank-cnn} are excluded since their dataset is different, as indicated by a higher 1-class accuracy of 79.8\%.
\begin{table}[h]
    \centering
    \caption{
        The effect of replacing ReLU or $\tanh$ with extrapolated Chebyshev-Lagrange (CL) activations on the DementiaBank healthy-versus-dementia classification scores (\%). Dotted lines separate the controls from the experimental condition for each architecture class. The parameter counts of width 32 and 64 ReLU/$\tanh$ are 6.5K and 21K respectively, and increase to 7.0K and 22K for CL. 1-class represents a model that predicts the majority class (i.e. healthy condition) only. Results represent means and standard deviations of 30 random initializations and sets of 10-fold cross validation. Lilliefors test for diagnosing non-normal distribution of the $\tanh$ and \textit{CL-extrapolate} results returned $p > 0.10$ for either distributions, validating the Student's $t$-test with no assumption of equal variance to compute the significance in their differences of means. $^{**}$ denotes $p < 0.01$ and $^{***}$ denotes $p < 0.001$. 
    }
    \label{chebyeffect-dementiabank}
    \begin{tabular}{l|c|c|c|c|c}
    \hline
     Activation & Width & Accuracy & Sensitivity & Specificity & Micro-F1 \\
     \hline
      1-class &  \multirow{4}{*}{32} & $56.3\pm  0.0$ & $0.0\pm0.0$ & $100.0\pm0.0$ & $0.0\pm0.0$\\
      ReLU &  & $82.5\pm1.2$ & $79.4\pm1.5$ & $84.9\pm1.8$ & $79.9\pm1.3$\\
      $\tanh$ &  & $82.7\pm1.2$ & $79.1\pm1.3$ & $85.4\pm1.5$ & $80.0\pm1.3$\\
      \hdashline
      \textit{CL-extrapolate} & & $\mathbf{84.7\pm 0.8^{***}}$ & $\mathbf{80.7\pm 1.2^{***}}$ & $87.7\pm1.0^{***}$ & $\mathbf{82.1\pm0.9^{***}}$\\
      \hline
      1-class &  \multirow{4}{*}{64} & $56.3\pm  0.0$ & $0.0\pm0.0$ & $100.0\pm0.0$ & $0.0\pm0.0$ \\
      ReLU &  & $82.9\pm1.0$ & $79.6\pm1.4$ & $85.4\pm1.2$ & $80.3\pm1.2$ \\
      $\tanh$ &  & $83.1\pm1.1$ & $79.7\pm1.8$ & $85.7\pm1.3$ & $80.5\pm1.3$\\
      \hdashline
      \textit{CL-extrapolate} & & $\mathbf{84.6\pm 0.9^{***}}$ & $\mathbf{80.7\pm 1.3^{**}}$ & $87.6\pm1.0^{***}$ & $\mathbf{82.0\pm1.1^{***}}$\\
      \hline
    \end{tabular}
\end{table}
\begin{table}[h]
    \caption{
        A comparison of 32-width, 2-block, 6-layer residual networks that use extrapolated Chebyshev-Lagrange (CL-ex) activations with previously reported results on DementiaBank healthy-versus-dementia classification scores (\%). CL-ex results for accuracy, micro-F1 and macro-F1 show means and standard deviations of 30 random initializations and sets of 10-fold cross validation. As per \cite{multiview-emb}, we also show the best macro-F1 across 10 validation folds for a single seed.
    }
    \label{dementiabank-globalcomparison}
    \centering
    \begin{tabular}{l|c|c|c|c}
        \hline
        Model & Accuracy & Micro-F1 & Macro-F1 & Best macro-F1 \\
        \hline
        1-layer neural net \cite{heterogeneousdata} & -- & $75.9$ & -- & -- \\ 
        TCN \cite{transductiveconsensus} & $75-77$ & -- & -- & --\\
        Random forest (RF) \cite{normativedata} & -- & $79.6$ & -- & --\\
        SVM \cite{normativedata} & -- & $80.6$ & -- & -- \\
        Multiview embed. + RF \cite{multiview-emb} & -- & -- & -- & $82.4\pm5.2$ \\
        Custom features + CL-ex ResNet & $84.7\pm 0.8$ & $82.1\pm0.9$ & $81.8\pm1.1$ & $84.0\pm7.5$ \\
        \hline
    \end{tabular}
\end{table}

\subsection{Image classification}
Table \ref{image-results} shows MNIST and CIFAR-10 results when we replace ReLU with \textit{CL-extrapolate} in the last 8 convolutional layers of a 14 layer residual network. No significant change is seen.
\begin{table}[h]
    \centering
    \caption{Results for image classification. Accuracy shows mean and standard deviations for 3 random initializations. Since normality of the two distributions is unknown, we use the Kruskal-Wallis H-test on the results of extrapolated Chebyshev-Lagrange versus ReLU to determine the p-value. }
    \label{image-results}
    \subfloat[MNIST]{
        \begin{tabular}{c|c|c}
            \hline
            Model & Accuracy & P-value \\
            \hline
            ReLU & $99.69\pm0.04$ & \multirow{2}{*}{0.38} \\
            \textit{CL-extrapolate} & $99.66\pm0.03$ & \\
            \hline
        \end{tabular}
        \label{mnist-results}
    }\quad
    \subfloat[CIFAR-10]{
        \begin{tabular}{c|c|c}
            \hline
            Model & Accuracy & P-value \\
            \hline
            ReLU & $93.45\pm0.10$ & \multirow{2}{*}{0.83} \\
            \textit{CL-extrapolate} & $93.40\pm0.29$ & \\
            \hline
        \end{tabular}
        \label{cifar-results}
    }
\end{table}
\begin{figure}[h]
    \begin{minipage}[t]{0.33\textwidth}\vspace{0pt}
        Figure \ref{hist-cifar} shows the distribution of inputs and the shape of select activations in the trained model. Similar results are seen for both MNIST and CIFAR-10. Appendix \ref{appendix-relu-vis} shows corresponding plots of the original ReLU network for reference. The distribution of inputs for ReLU or \textit{CL-extrapolate} appear to be bell-shaped and centered close to the origin for the majority of units, although there is also a minority of \textit{CL-extrapolate} inputs which form bi-modal distributions. The ranges of inputs appear consistent between the two activations. The typical hidden unit of shallow layers cover intervals in $\pm10$ to $\pm30$, while deeper layers cover intervals around $\pm5$.
        Since the range of input for shallow layers is significantly greater than the range covered by the polynomial, we observe \textit{CL-extrapolate} adopting shapes that emphasize piece-wise linearity. Figure \ref{hist-shallow} shows \textit{CL-extrapolate} learning various configurations of V-shape or PReLU activations. The densest part of the distribution is still within the polynomial region. For deeper layers, we see greater use of the polynomial region and smoother outputs, as illustrated in Figure \ref{hist-deep}.
    
    \strut\end{minipage}\hspace{.02\textwidth}
    \begin{minipage}[t]{0.65\textwidth}\vspace{0pt}
        \subfloat[Activations in the 6th convolutional layer.]{
            \includegraphics[scale=0.21]{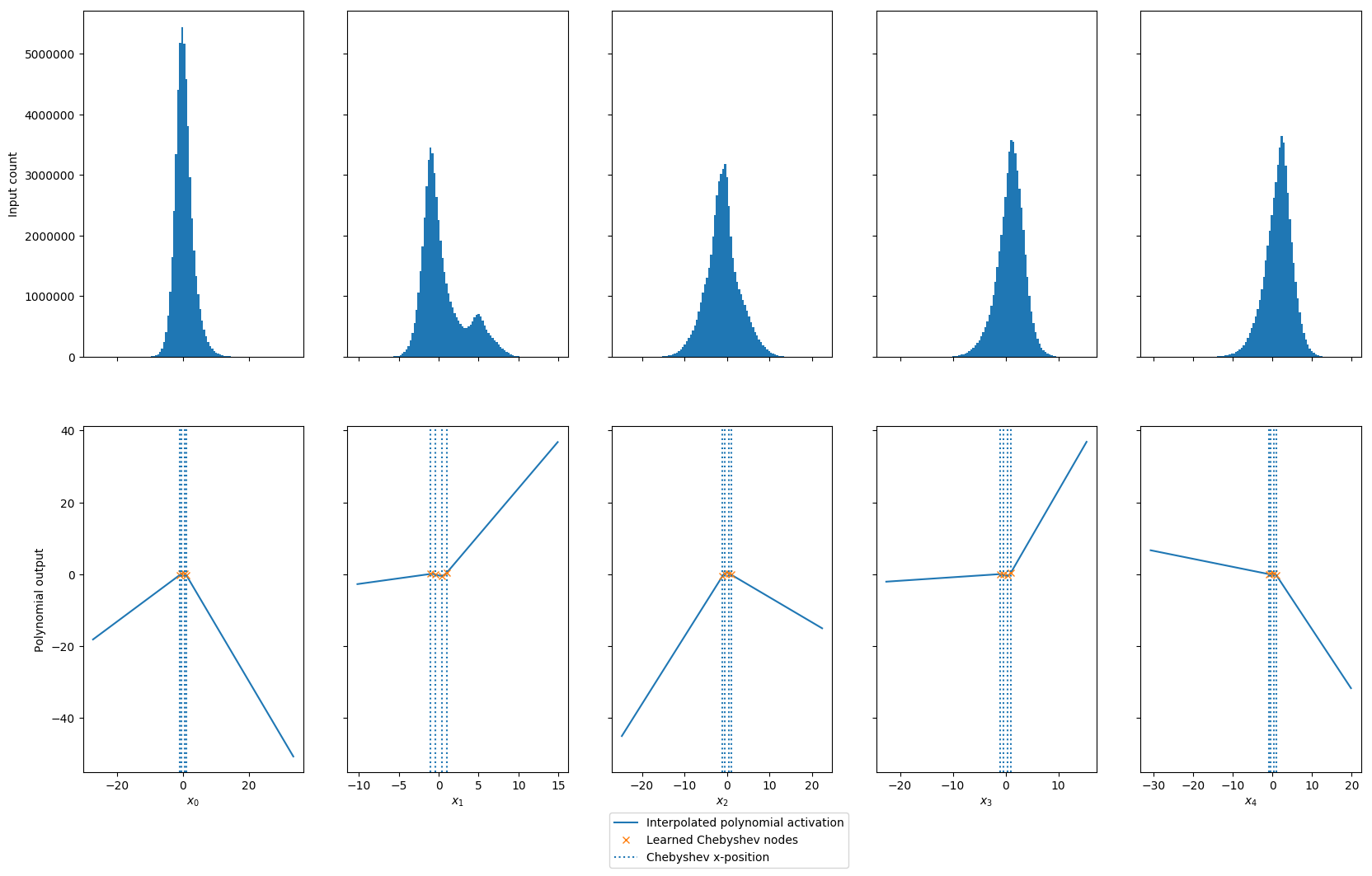}
            \label{hist-shallow}
        }\\
        \subfloat[Activations in the 13th convolutional layer.]{
            \includegraphics[scale=0.21]{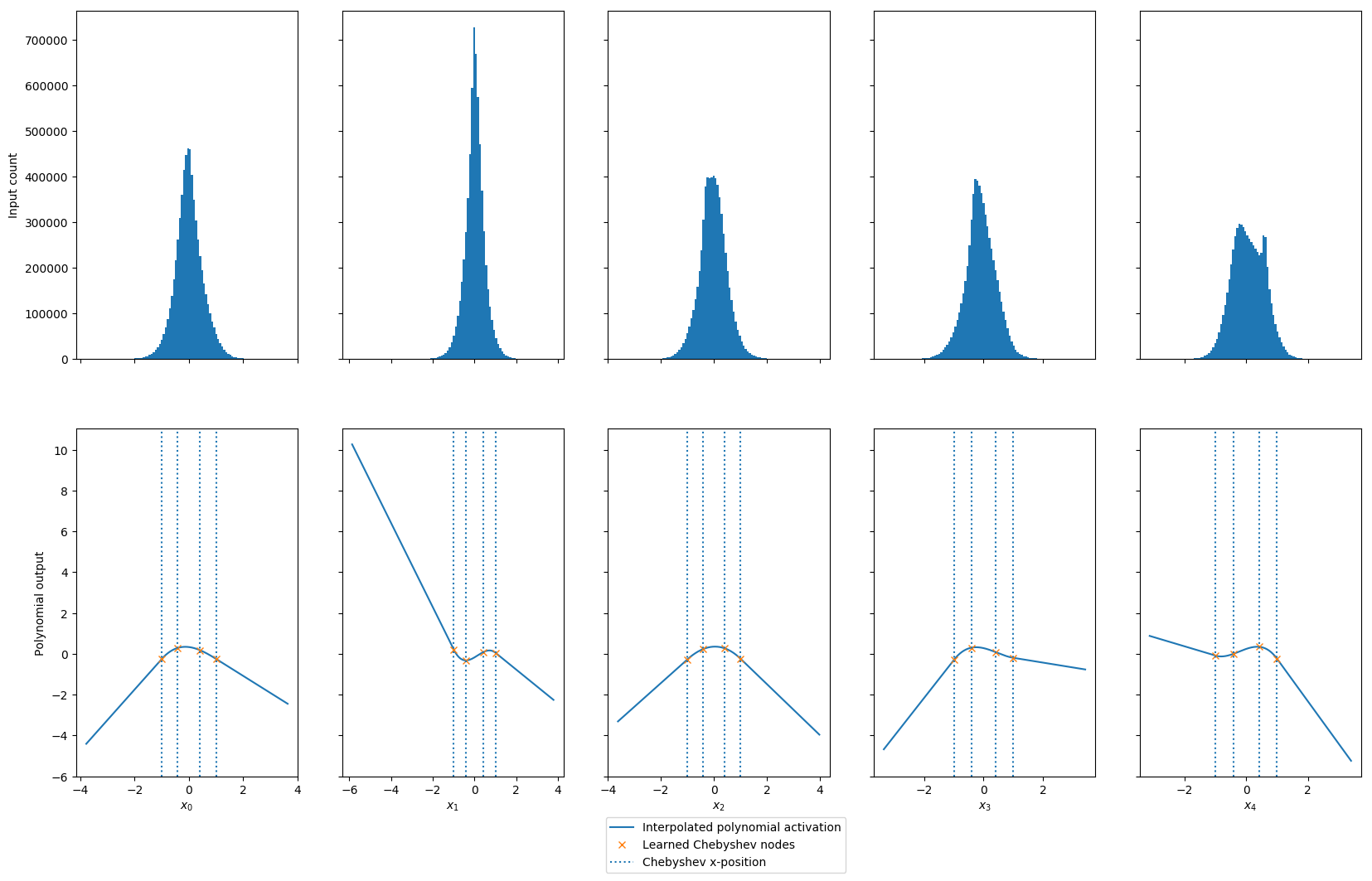}
            \label{hist-deep}
        }
        \caption{
            Sample plots of activations and histograms of their inputs for the first 5 hidden units at the 6th (\ref{hist-shallow}) and 13th (\ref{hist-deep}) layers of the model trained on CIFAR-10. The $x$-axis range shows the minimum and maximum input value per element.
        }
        \label{hist-cifar}
    \end{minipage}
\end{figure}

\section{Discussion}
\label{discussion}
The identical performance of the models used in image classification does not seem to be caused by \textit{CL-extrapolate} behaving identically to ReLU, as shown in Figure \ref{hist-cifar}. We can interpret the mechanism of ReLU as reporting the one-sided variance of its input distribution (Appendix \ref{appendix-relu-vis}). On the other hand, \textit{CL-extrapolate} can learn to report one or two-sided variances, induce and report bimodal distributions from its input network, and produce smoothly interpolated outputs (Figure \ref{hist-cifar}). These properties allow for models to adapt to both smooth and non-smooth functions, which may be the reason for their superior performance over similar variants or controls on the synthetic dataset experiments (Table \ref{synthetic-0.01}). Consistently low RMSE on the various non-linear relationships of the synthetic datasets indicates that \textit{CL-extrapolate} reduces overfitting and learns the true relationship between inputs and outputs better than the other tested activations. This implies that models can improve in data efficiency by replacing their activation functions with \textit{CL-extrapolate}. We replace ReLU or $\tanh$ with \textit{CL-extrapolate} in two different architectures trained on the task of classifying dementia diagnosis from speech and observe significantly closer performance of these models to the diagnoses of real physicians (Table \ref{chebyeffect-dementiabank}). These benefits come for free in respect to the acquisition of data, which may be of use for other machine learning tasks where data is slow or difficult to obtain.

Given this, we consider why \textit{CL-extrapolate} has limited effect on image classification (Table \ref{image-results}). One likely explanation is that the combination of noise level and function characteristic in image classification negates the interpolation benefits of \textit{CL-extrapolate}. Lagrangian interpolation of Chebyshev nodes minimizes the error bound between the true, smooth function and the polynomial of noise-free observations. When the true function is not smooth, the error bound described in equation \ref{chebyerror} no longer holds as $\max_{\xi\in[-1,1]}\left|f^{(n)}(\xi)\right| $ is not defined. However, we also observed that models that use \textit{CL-extrapolate} continue to enjoy superior interpolation even on non-smooth datasets (Table \ref{synthetic-0.01}). We observe reduced impact of \textit{CL-extrapolate} on non-smooth function modelling only with higher levels of noise contamination (Appendix \ref{appendix-samplesynthetic}, Table \ref{synthetic-0.04}). This suggests that we should use \textit{CL-extrapolate} when the input-output relationship is suspected to be smooth or the data has low levels of noise. Other options include finding ways of inducing the input and output into an artificial, smooth relationship, or applying algorithms that reduce noise, such as employing feature selection as we did in the DementiaBank experiment (Section \ref{methods-dementiabank}). An interesting experiment would be to use \textit{CL-extrapolate} to map intermediate features of deep models trained on large datasets to target outputs, assuming that these features have reduced noise compared to the raw input.

Costs for using \textit{CL-extrapolate} include increased memory and computation complexity. The implementation and gradient equations of \textit{CL-extrapolate} in Appendices \ref{appendix-lagrangecalc} and \ref{appendix-lagrangegrad} clearly show respective complexities of $O(n^2)$ and $O(n^3)$ of the hyperparameter choice $n$ for the maximum polynomial degree, which is multiplied to the base complexity of the network. We find $n=3$ to be sufficient for our experiments, but the current implementation may be impractical for the lowest convolutional layers of modern image classifiers or with much larger choices of $n$ (i.e., $n=100,1000...$).

Future work should examine the theoretical properties of these activations on generalizing the outcomes of the synthetic experiments. In particular, it is not clear why \textit{CL-extrapolate} showed superior performance on non-smooth datasets with low noise, or if there exist sets of problems where certain CL variants are likely to consistently out-perform ReLU or $\tanh$. Further research should investigate how noise influences the degradation of performance for these activations and regularization techniques to reduce the degradation rate.

\section{Conclusion}

We implement a new piecewise continuous activation function composed of the Lagrangian interpolation of parameterized Chebyshev nodes on $[-1, 1]$ and extrapolated linear components outside of this range. We observe consistent and better interpolation results compared to ReLU, $\tanh$ and other activation functions on a variety of smooth and non-smooth synthetic experiments. Significant improvement is seen in DementiaBank, at no cost to MNIST or CIFAR-10 classification. We show how the proposed activations are more versatile than ReLU, which appear to explain their superior performance. 

\clearpage

\bibliographystyle{ieee}
\bibliography{main}

\clearpage
\section{Appendix}

\subsection{Parameters}
\label{appendix-chebyparams}

We first consider the case where the input to the activation function is a vector $v$ of length $d$, with elements $v_i,\: i=1,...,d$. 
For the hyperparameter $n$, we wish to learn a separate $n$-degree polynomial for every element $v_i$ that minimizes the output error of the network.
Since we choose to use Lagrangian polynomial interpolation, we need a total of $n+1$ nodes for each element.
We fix the $x$-coordinates of the $n+1$ nodes across all features to be the scaled Chebyshev nodes with radius $r$:

$$x_k = r \cos(\frac{2k-1}{2(n+1)}\pi),\: k=1,...,n+1$$

To ensure $x_1=1$ and $x_{n+1}=-1$, we choose the radius to be:

$$r =  \cos(\frac{\pi}{2(n+1)})^{-1}$$

Therefore we only have to learn the $y$-coordinates of the nodes, which amounts to a total of $(n+1)\times d$ parameters and can be represented by a matrix.

Next, we extend the implementation of the activation to tensors with dimensions greater than 1. 
Suppose the input $V$ is a tensor of shape $W\times H\times d$, where $d=3$ for an RGB image.
We can consider the channel vectors of this tensor (i.e. the vectors of length $d$ on the $W\times H$ plane) as the same type of vector input $v$ as described above. In a similar way, any high dimensional tensor can be interpreted as feature vectors placed in the remaining shape.
Therefore the number of parameters for higher dimensional tensors is the same as the single vector case.

In the following sub-section, we describe the computation of the Lagrangian polynomial that results from these nodes.

\subsection{Lagrangian polynomial calculation}
\label{appendix-lagrangecalc}

The Lagrangian polynomial $P_n$ with degree $n$ that arises from $n+1$ nodes with parameterized $y$-positions is described by

$$ P_n(v_i) = \sum_{j=1}^{n+1}y_j \ell_j(v_i),
\;
\text{where}
\;
\ell_j(v_i) := \prod_{\substack{1\le m\le n+1\\ m\ne j}}\frac{v_i-x_m}{x_j-x_m}
\;
\text{are the Lagrangian basis functions.}
$$

Since the basis functions only rely on $x$-coordinates, we can represent the numerator operation as a matrix of shape $d\times n$ and precompute the denominator into a vector of shape $d$.
We program Algorithm \ref{lagrange-compute} in a few lines of PyTorch and will release our code.

\begin{algorithm}
    \caption{Computation of Lagrange polynomials from Chebyshev nodes}
    \label{lagrange-compute}
    \begin{algorithmic}[1]
        \Notation{
            $I$ denotes the identity matrix \\
            $[\cdot]$ denotes element-wise index operation \\
            $\prod$ denotes the product of elements along the last axis of the tensor \\
            $\sum$ denotes the sum of elements in the vector
        }
        \Inputs{
            $x$ is a vector of Chebyshev $x$-coordinates of length $n+1$ \\
            $y$ is a vector of $y$-coordinates for each $x_i$, of length $n+1$ \\
            $v_i$ is a real number representing the $i$th input feature
        }
        \Cache{
            $X \leftarrow$ repeatedly stack $x$ into a square matrix \Comment{shape $(n+1)^2$} \\
            numerator $\leftarrow X[\sim I]$ \Comment{shape $(n+1)\times n$}\\
            denominator $\leftarrow \prod (x^T - \text{numerator})$ \Comment{shape $n+1$}
        }
        \State{
            $\ell \leftarrow \prod(v_i-\text{numerator})/\text{denominator}$ \Comment{shape $n+1$}
        }
        \State{
            \Return{$y\cdot\ell$} \Comment{shape of 1, polynomial output from $v_i$}
        }
    \end{algorithmic}
\end{algorithm}

\subsection{Pre-computation of gradients for Lagrangian bases}
\label{appendix-lagrangegrad}

Recall the gradient of $P_n$:
    
$$
P_n'(x) = \sum^{n+1}_{j=1}y_j\ell_j'(x)
$$

where

$$
\ell_j'(x) := \sum^{n+1}_{i=1,i\neq j}\frac{1}{x_j-x_i}\prod^{n+1}_{m=1, m\ne (i, j)}\frac{x-x_m}{x_j-x_m}
$$

The similarity between the Lagrangian polynomial and its gradient allows for the reuse of outputs from Algorithm \ref{lagrange-compute}, as shown in Algorithm \ref{lagrange-grad}. We can use this to quickly compute $P_n'(-1)$ and $P_n'(+1)$  for linear extrapolation.

\begin{algorithm}
    \caption{Computation of Lagrange gradients}
    \label{lagrange-grad}
    \begin{algorithmic}[1]
        \Notation{
            $I$ denotes the identity matrix \\
            $[:,\cdot]$ denotes element-wise index operation along the last axis\\
            $\prod$ denotes the product of elements along the last axis of the tensor \\
            $\sum$ denotes the sum of elements along the last axis of the tensor
        }
        \Inputs{
            numerator is the cached matrix of shape $(n+1)\times n$ in algorithm \ref{lagrange-compute} \\
            denominator is a cached vector of shape $(n+1)$ in algorithm \ref{lagrange-compute} \\
            $y$ is a vector of $y$-coordinates for each $x_i$, of length $n+1$ \\
            $c$ is a constant real number at which to compute the gradient of the polynomial
        }
        \Cache{
            square\_numerator $\leftarrow$ repeatedly stack numerator on the last axis into a square \\
            \Comment{shape $(n+1)\times n^2$}\\
            numerator\_grad $\leftarrow$ square\_numerator$[:, \sim I]$ \Comment{shape $(n+1)\times n\times (n-1)$} \\
            right\_prod $\leftarrow\prod(c-\text{numerator\_grad})$ \Comment{shape $(n+1)\times n$} \\
            $\ell' \leftarrow \sum(\text{right\_prod}/\text{denominator})$ \Comment{shape $n+1$}
        }
        \State{
            \Return{$y\cdot\ell'$} \Comment{shape of 1, gradient at $c$}
        }
    \end{algorithmic}
\end{algorithm}

\clearpage
\subsection{Features used in DementiaBank classification}
\label{appendix-dementiabank-borutafeatures}

We use Boruta feature selection on a different but related dataset, FamousPeople, to obtain the set of 66 relevant features from the original full set of 480 features (Table \ref{appendix-db-borutafeatures-description}). 

\begin{table}[h]
    \centering
    \caption{The 66 Boruta-selected feature used in DementiaBank classification. They are described in \cite{heterogeneousdata}.}
    \label{appendix-db-borutafeatures-description}
    \begin{tabular}{c|c|c|c}
        \hline
        Number & Feature & Number & Feature \\
        \hline
        1 & long\_pause\_count\_normalized & 34 & graph\_lsc\\
        2 & medium\_pause\_duration & 35 & graph\_density\\
        3 & mfcc\_skewness\_13 & 36 & graph\_num\_nodes\\
        4 & mfcc\_var\_32 & 37 & graph\_asp\\
        5 & TTR & 38 & graph\_pe\_undirected\\
        6 & honore & 39 & graph\_diameter\_undirected\\
        7 & tag\_IN & 40 & Lu\_DC\\
        8 & tag\_NN & 41 & Lu\_DC/T\\
        9 & tag\_POS & 42 & Lu\_CT\\
        10 & tag\_VBD & 43 & local\_coherence\_Google\_300\_avg\_dist\\
        11 & tag\_VBG & 44 & local\_coherence\_Google\_300\_min\_dist\\
        12 & tag\_VBZ & 45 & constituency\_NP\_type\_rate\\
        13 & pos\_NOUN & 46 & constituency\_PP\_type\_prop\\
        14 & pos\_ADP & 47 & constituency\_PP\_type\_rate\\
        15 & pos\_VERB & 48 & ADJP\_->\_JJ\\
        16 & pos\_ADJ & 49 & ADVP\_->\_RB\\
        17 & category\_inflected\_verbs & 50 & NP\_->\_DT\_NN\\
        18 & prp\_ratio & 51 & NP\_->\_PRP\\
        19 & nv\_ratio & 52 & PP\_->\_TO\_NP\\
        20 & noun\_ratio & 53 & ROOT\_->\_S\\
        21 & speech\_rate & 54 & SBAR\_->\_S\\
        22 & avg\_word\_duration & 55 & VP\_->\_VB\_ADJP\\
        23 & age\_of\_acquisition & 56 & VP\_->\_VBG\\
        24 & NOUN\_age\_of\_acquisition & 57 & VP\_->\_VBG\_PP\\
        25 & familiarity & 58 & VP\_->\_VBG\_S\\
        26 & VERB\_frequency & 59 & VP\_->\_VBZ\_VP\\
        27 & imageability & 60 & avg\_word\_length\\
        28 & NOUN\_imageability & 61 & info\_units\_bool\_count\_object\\
        29 & sentiment\_arousal & 62 & info\_units\_bool\_count\_subject\\
        30 & sentiment\_dominance & 63 & info\_units\_bool\_count\\
        31 & sentiment\_valence & 64 & info\_units\_count\_object\\
        32 & graph\_avg\_total\_degree & 65 & info\_units\_count\_subject\\
        33 & graph\_pe\_directed & 66 & info\_units\_count\\
        \hline
    \end{tabular}
    
\end{table}

\clearpage
\subsection{Sample visualizations of the synthetic datasets}
\label{appendix-samplesynthetic}

Figures \ref{appendix-samplesynthetic-smooth} and \ref{appendix-samplesynthetic-nonsmooth} provide sample visualizations of the synthetic datasets by varying $x_0$ from $-1$ to $+1$ while fixing $x_i=0.5$ for $i>0$. We choose $0.5$ instead of $0$ because some of the functions are produced from the multiplication of inputs and we wanted to see if the networks could model more interesting dynamics. Note that the RMSE of the figures correspond to the test result, which randomly samples all $x_i$ from the uniform distribution on $[-1, 1]$, and do not correspond to the RMSE of the visualization.

\begin{table}[h]
    \centering
    \caption{Randomly selected visualizations of the smooth synthetic datasets. These are produced by varying $x_0$ from $-1$ to $+1$ while fixing $x_i=0.5$ for $i>0$. We compare the interpolation quality of models that use $\tanh$ versus the extrapolated Chebyshev-Lagrange variant (\textit{CL-extrapolate}).}
    \label{appendix-samplesynthetic-smooth}
    \begin{tabular}{ccc}
        \hline
         Dataset & $\tanh$ &\textit{CL-extrapolate} \\
         \hline
         Pendulum & \includegraphics[valign=c,scale=0.35]{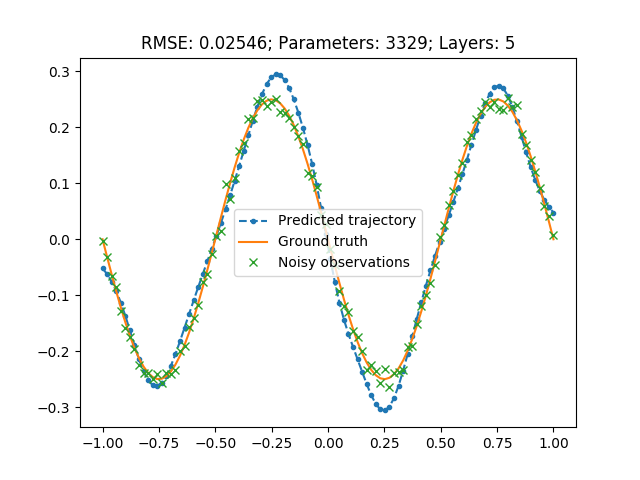} & \includegraphics[valign=c,scale=0.35]{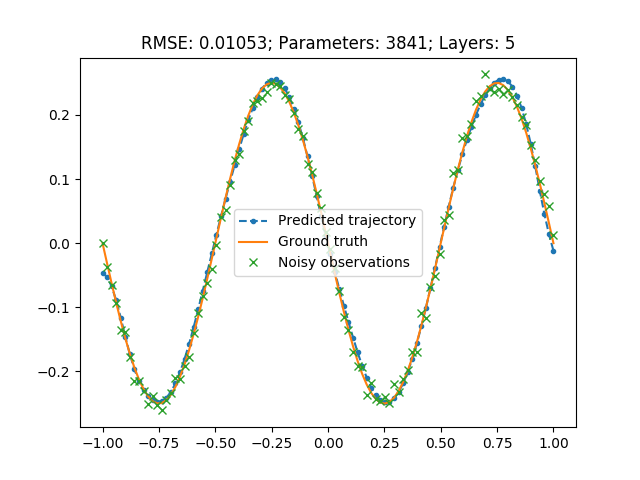} \\
         Arrhenius & \includegraphics[valign=c,scale=0.35]{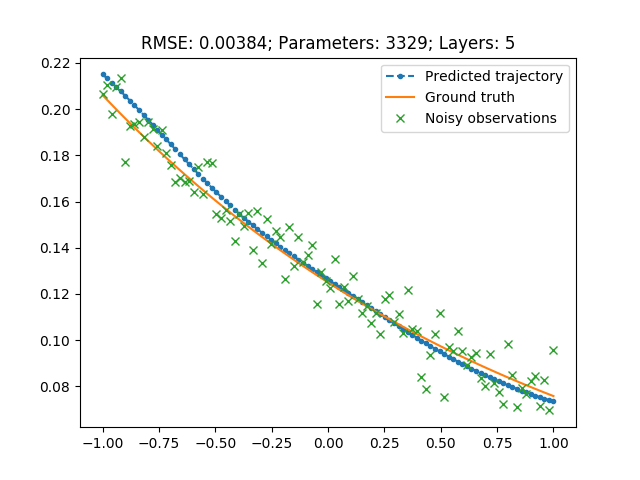} & \includegraphics[valign=c,scale=0.35]{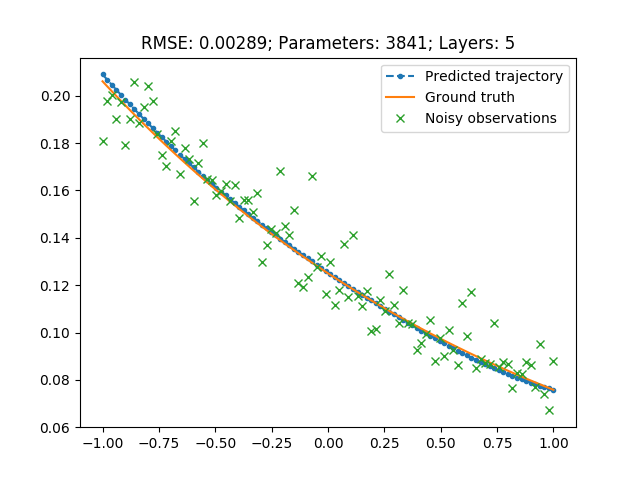} \\
         Gravity & \includegraphics[valign=c,scale=0.35]{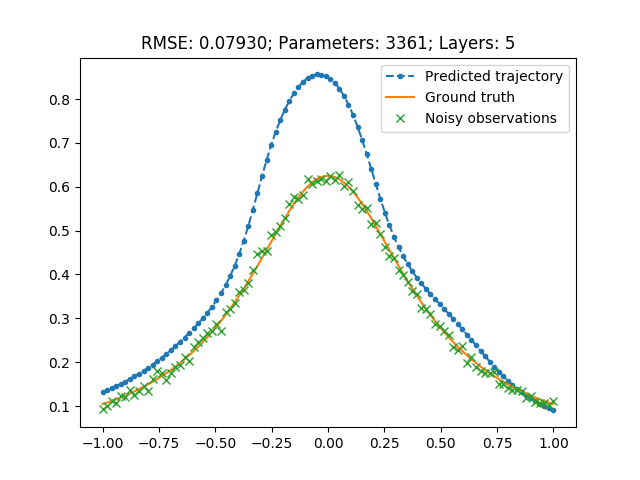} & \includegraphics[valign=c,scale=0.35]{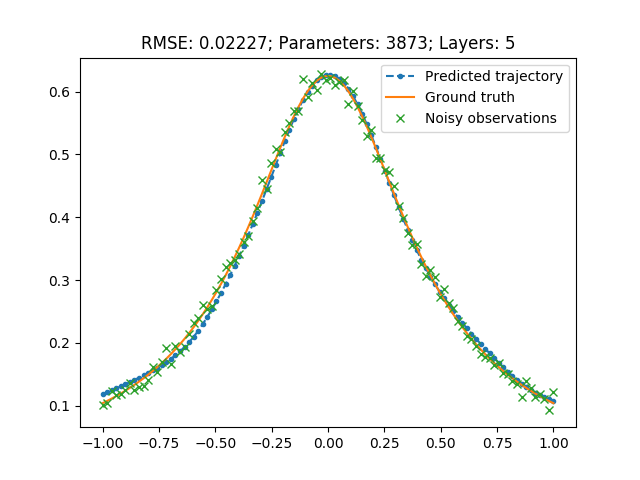} \\
         Sigmoid & \includegraphics[valign=c,scale=0.35]{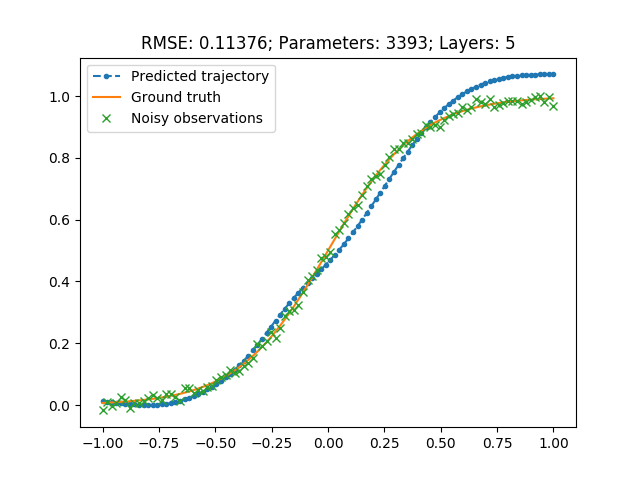} & \includegraphics[valign=c,scale=0.35]{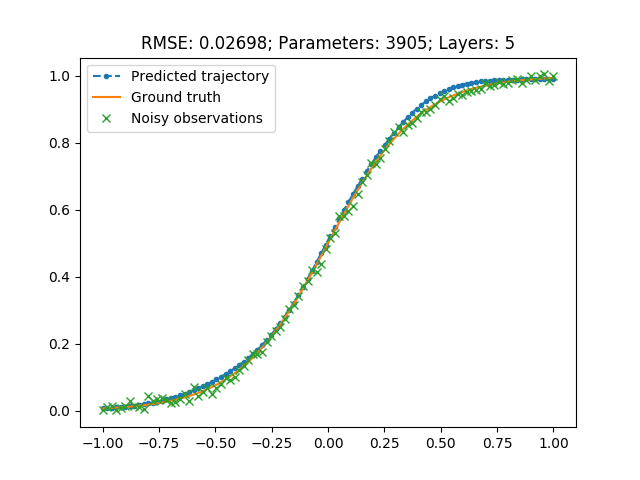} \\
         \hline
    \end{tabular}
\end{table}

\begin{table}[h]
    \centering
    \caption{Randomly selected visualizations of the non-smooth synthetic datasets. These are produced by varying $x_0$ from $-1$ to $+1$ while fixing $x_i=0.5$ for $i>0$. We compare the interpolation quality of models that use ReLU versus the extrapolated Chebyshev-Lagrange variant (\textit{CL-extrapolate}).}
    \label{appendix-samplesynthetic-nonsmooth}
    \begin{tabular}{ccc}
        \hline
         Dataset & ReLU &\textit{CL-extrapolate} \\
         \hline
         Jump & \includegraphics[valign=c,scale=0.35]{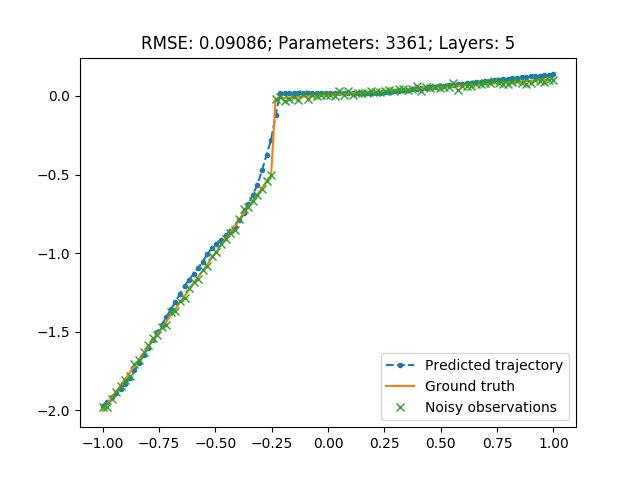} & \includegraphics[valign=c,scale=0.35]{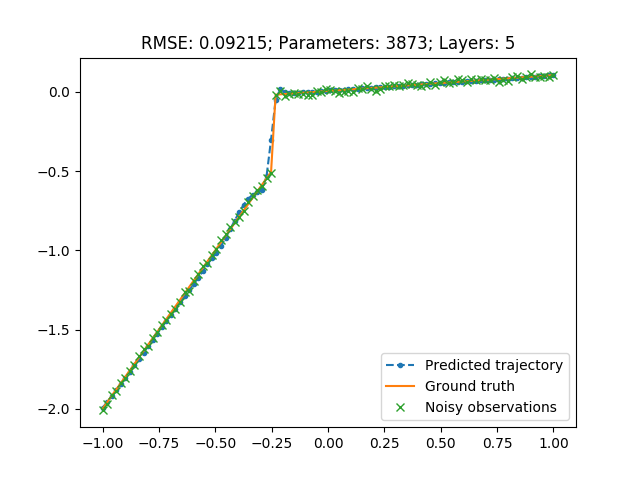} \\
         PReLU & \includegraphics[valign=c,scale=0.35]{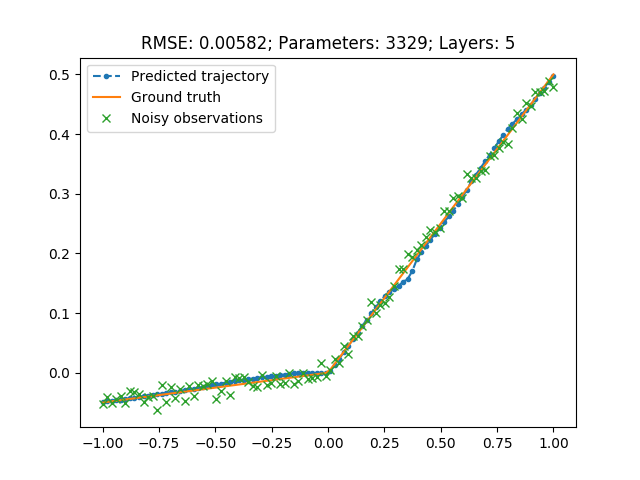} & \includegraphics[valign=c,scale=0.35]{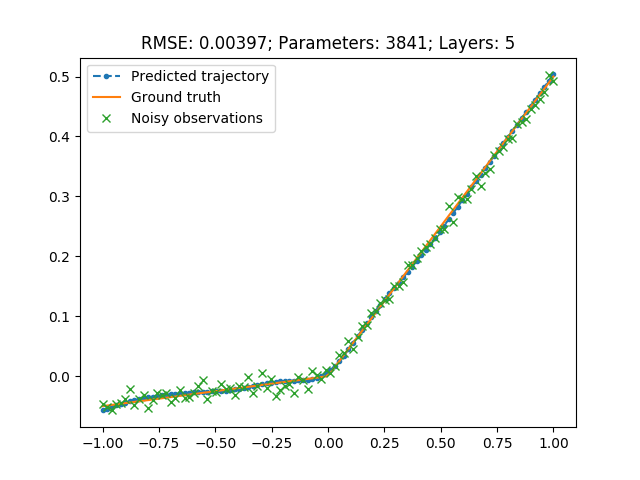} \\
         Step & \includegraphics[valign=c,scale=0.35]{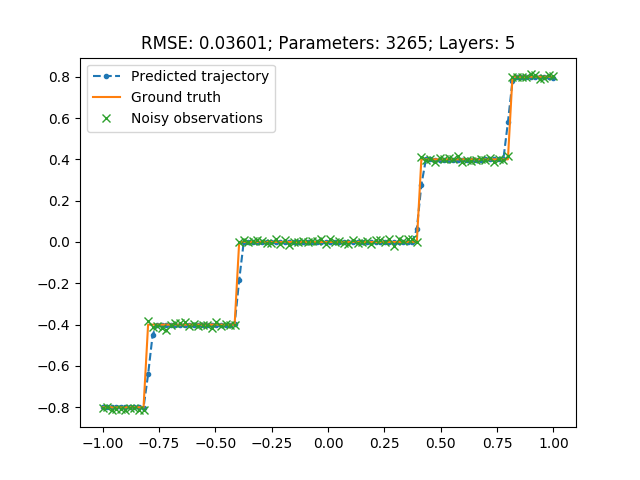} & \includegraphics[valign=c,scale=0.35]{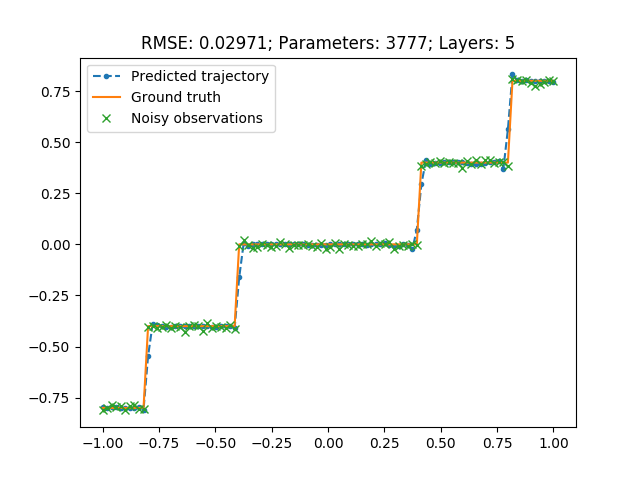} \\
         \hline
    \end{tabular}
\end{table}

\clearpage
\subsection{Results on the synthetic datasets with increased Gaussian noise}
\label{appendix-syntheticnoisy}

Table \ref{synthetic-0.04-samples} shows sample visualizations of the magnitude of noise. Table \ref{synthetic-0.04} shows the results of the CL variants on the synthetic datasets with greater levels of Gaussian noise (i.e. 0.04 standard deviations).

\begin{table}[h]
    \centering
    \caption{Randomly selected visualizations of some synthetic datasets with Gaussian noise of 0.04 standard deviations. These are produced by varying $x_0$ from $-1$ to $+1$ while fixing $x_i=0.5$ for $i>0$. These samples are from a model that uses extrapolated Chebyshev-Lagrange.}
    \label{synthetic-0.04-samples}
    \begin{tabular}{cc}
        \hline
        Smooth & Non-smooth \\
        \hline
         \includegraphics[valign=c,scale=0.35]{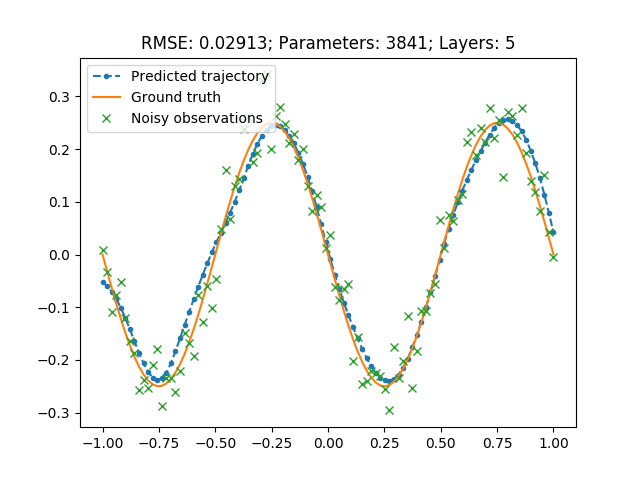} &
         \includegraphics[valign=c,scale=0.35]{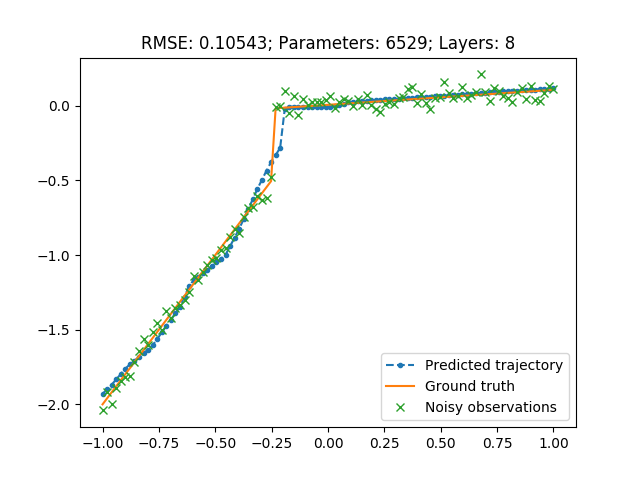} \\
         \includegraphics[valign=c,scale=0.35]{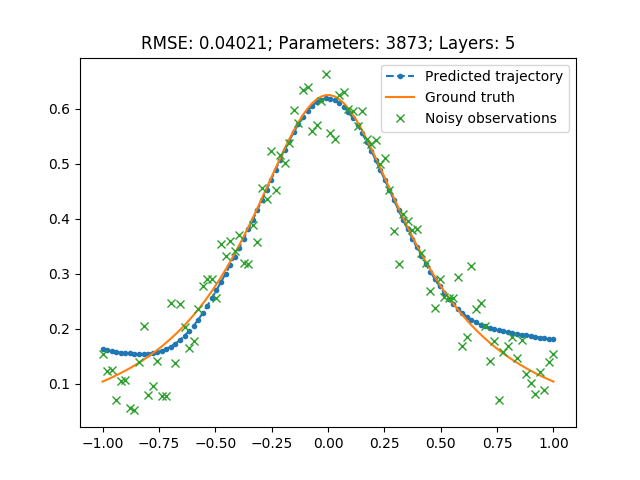} &
         \includegraphics[valign=c,scale=0.35]{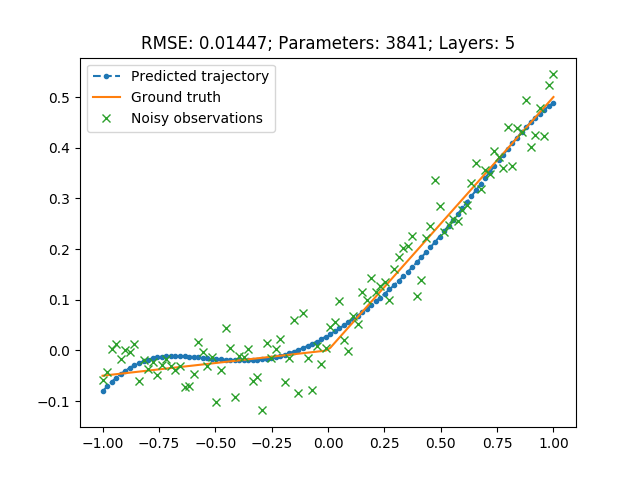} \\
         \includegraphics[valign=c,scale=0.35]{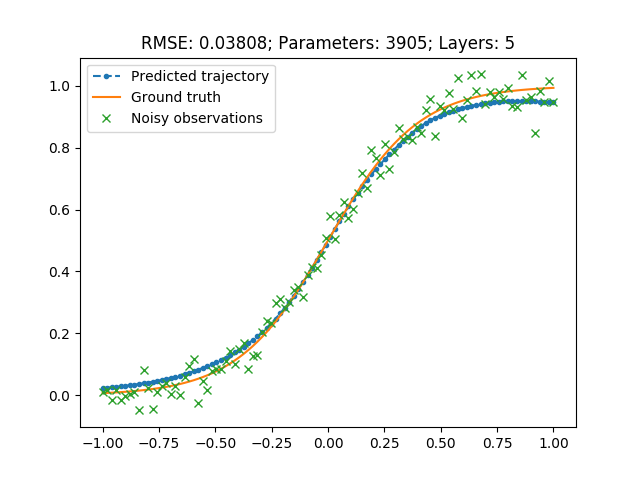} & \includegraphics[valign=c,scale=0.35]{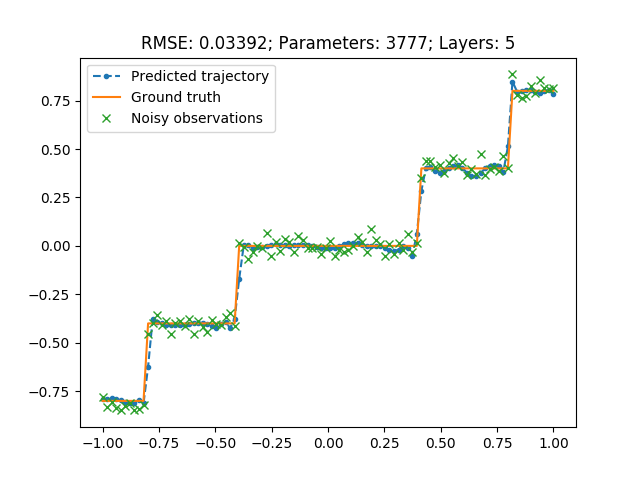} \\
         \hline
    \end{tabular}
\end{table}

\begin{table}[h]
    \centering
    \caption{Root mean squared error (RMSE) for the smooth (\ref{synthetic-smooth-0.04}) and non-smooth (\ref{synthetic-nonsmooth-0.04}) synthetic datasets with 0.04 standard deviations of Gaussian noise. NaN indicates counts of gradient explosion. See table \ref{synthetic-description} and \ref{synthetic-architectures} for descriptions of the datasets and models. Excluding rows containing any NaN, bold font marks the two most data-efficient methods while red font marks the least data-efficient methods.}
    \label{synthetic-0.04}
    \subfloat[Smooth datasets]{
        \label{synthetic-smooth-0.04}
        \begin{tabular}{c|c|c|c|c}
            \hline
            Model & Pendulum & Arrhenius & Gravity & Sigmoid \\
            \hline
            ReLU & $\badscore{0.140\pm0.045}$& $\badscore{0.0143\pm0.0016}$ & $\badscore{0.073\pm0.014}$ & $\badscore{0.069\pm0.007}$\\
            ReLU (2x depth) & $\badscore{0.204\pm0.020}$ & $\badscore{0.0142\pm0.0014}$ & $\badscore{0.202\pm0.165}$ & $\badscore{0.069\pm0.008}$\\
            ReLU (2x layers) & $\badscore{0.139\pm0.042}$ & $\badscore{0.0139\pm0.0010}$ & $\badscore{0.076\pm0.056}$ & $\badscore{0.052\pm0.005}$\\
            $\tanh$ & $0.049\pm0.020$ & $\mathbf{0.0088\pm0.0009}$ & $\badscore{0.073\pm0.053}$ & $\badscore{0.122\pm0.027}$\\
            Cubic & (10/10 NaN) & (4/10 NaN) & (10/10 NaN) & (10/10 NaN) \\
            \hline
            CL & (4/10 NaN) & $0.0082\pm0.0007$ & (10/10 NaN) & (10/10 NaN) \\
            WCP & (8/10 NaN) & $0.0091\pm0.0012$ & (10/10 NaN) & (10/10 NaN) \\
            \hline
            PCS-CL & $0.041\pm0.003$ & $0.0119\pm0.0009$ & $\mathbf{0.032\pm0.005}$ & $\mathbf{0.043\pm0.005}$\\
            $\tanh$-CL & $\badscore{0.191\pm0.048}$ & $0.0121\pm0.0018$ & $0.056\pm0.010$ & $0.049\pm0.012$\\
            \hline
            \textit{CL-regression} & $\mathbf{0.035\pm0.003}$ & $\badscore{0.0133\pm0.0012}$ & $0.062\pm0.006$ & $0.049\pm0.006$\\
            \textit{CL-extrapolate} & $\mathbf{0.028\pm0.001}$ & $\mathbf{0.0115\pm0.0011}$ & $\mathbf{0.040\pm0.003}$ & $\mathbf{0.040\pm0.004}$\\
            \hline
        \end{tabular}
    }
    
    \subfloat[Non-smooth datasets]{
        \label{synthetic-nonsmooth-0.04}
        \begin{tabular}{c|c|c|c}
            \hline
            Model & Jump & PReLU & Step \\
            \hline
            ReLU & $\badscore{0.12\pm0.01}$ & $\badscore{0.0155\pm0.0009}$ & $\badscore{0.052\pm0.017}$\\
            ReLU (2x depth) & $\badscore{0.17\pm0.14}$ & $\badscore{0.0170\pm0.0040}$ & $\mathbf{0.033\pm0.009}$\\
            ReLU (2x layers) & $\badscore{0.11\pm0.02}$ & $\badscore{0.0159\pm0.0045}$ & $0.035\pm0.011$\\
            $\tanh$ & $\mathbf{0.10\pm0.03}$ & $\badscore{0.0155\pm0.0072}$ & $\badscore{0.054\pm0.014}$\\
            Cubic & (10/10 NaN) & (9/10 NaN) & (7/10 NaN) \\
            \hline
            CL & (10/10 NaN) & (1/10 NaN) & $0.099\pm0.020$\\
            WCP & (10/10 NaN) & (2/10 NaN) & $0.104\pm0.021$\\
            \hline
            PCS-CL & $\mathbf{0.10\pm0.02}$ & $\mathbf{0.0111\pm0.0008}$ & $\badscore{0.070\pm0.006}$\\
            $\tanh$-CL  & $\badscore{0.12\pm0.02}$ & $\mathbf{0.0116\pm0.0007}$ & $\badscore{0.091\pm0.016}$\\
            \hline
            \textit{CL-regression} & $\mathbf{0.08\pm0.02}$ & $0.0152\pm0.0015$ & $\mathbf{0.024\pm0.001}$\\
            \textit{CL-extrapolate} & $\mathbf{0.10\pm0.02}$ & $0.0142\pm0.0013$ & $0.034\pm0.002$\\
            \hline
        \end{tabular}
    }
    
\end{table}

\clearpage
\subsection{Sample visualizations of ReLU activation in CIFAR-10}
\label{appendix-relu-vis}

\begin{figure}[h]
    \centering
    \subfloat[Activations in the 6th convolutional layer.]{
            \includegraphics[scale=0.32]{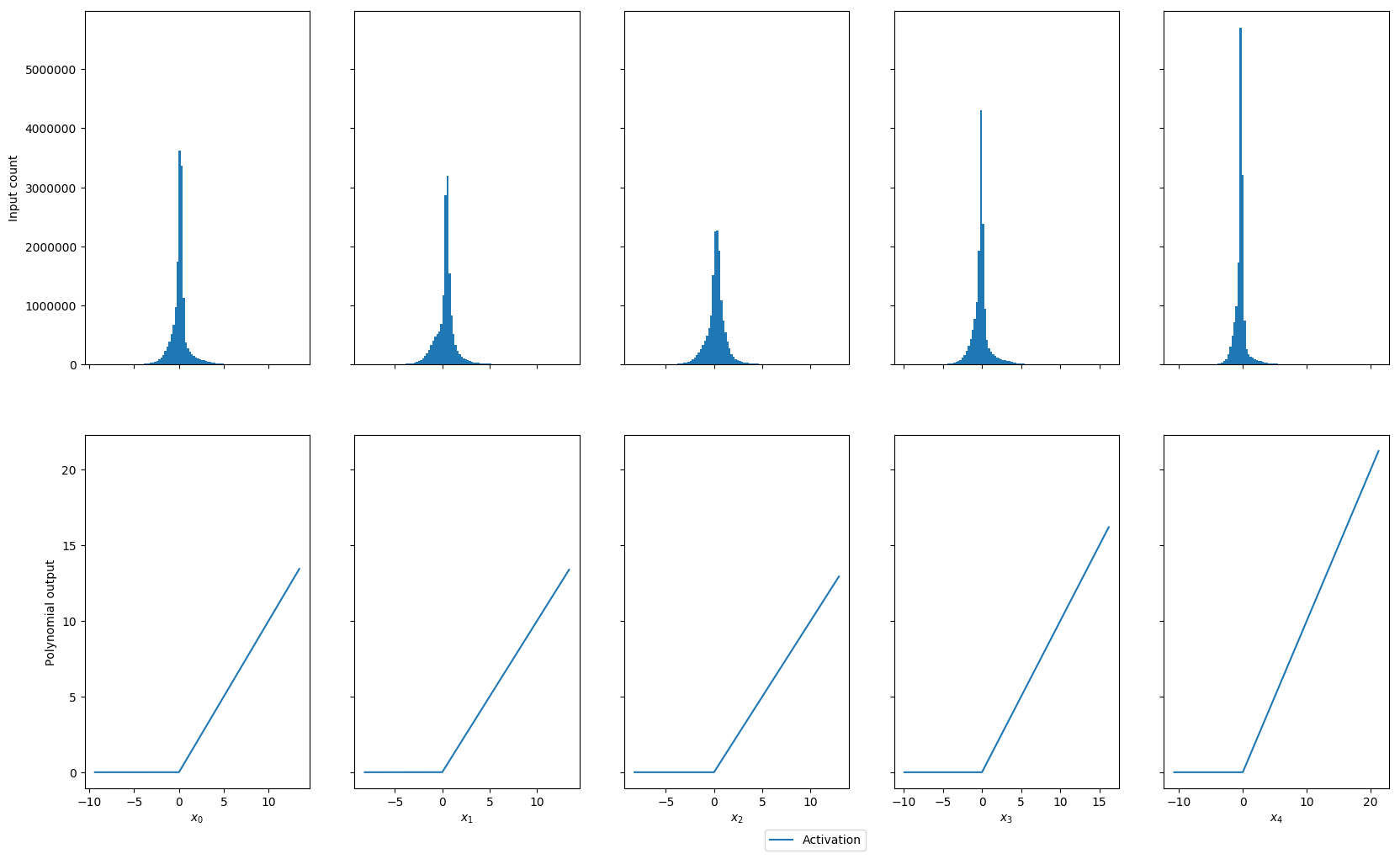}
            \label{relu-hist-shallow}
        }\\
        \subfloat[Activations in the 13th convolutional layer.]{
            \includegraphics[scale=0.32]{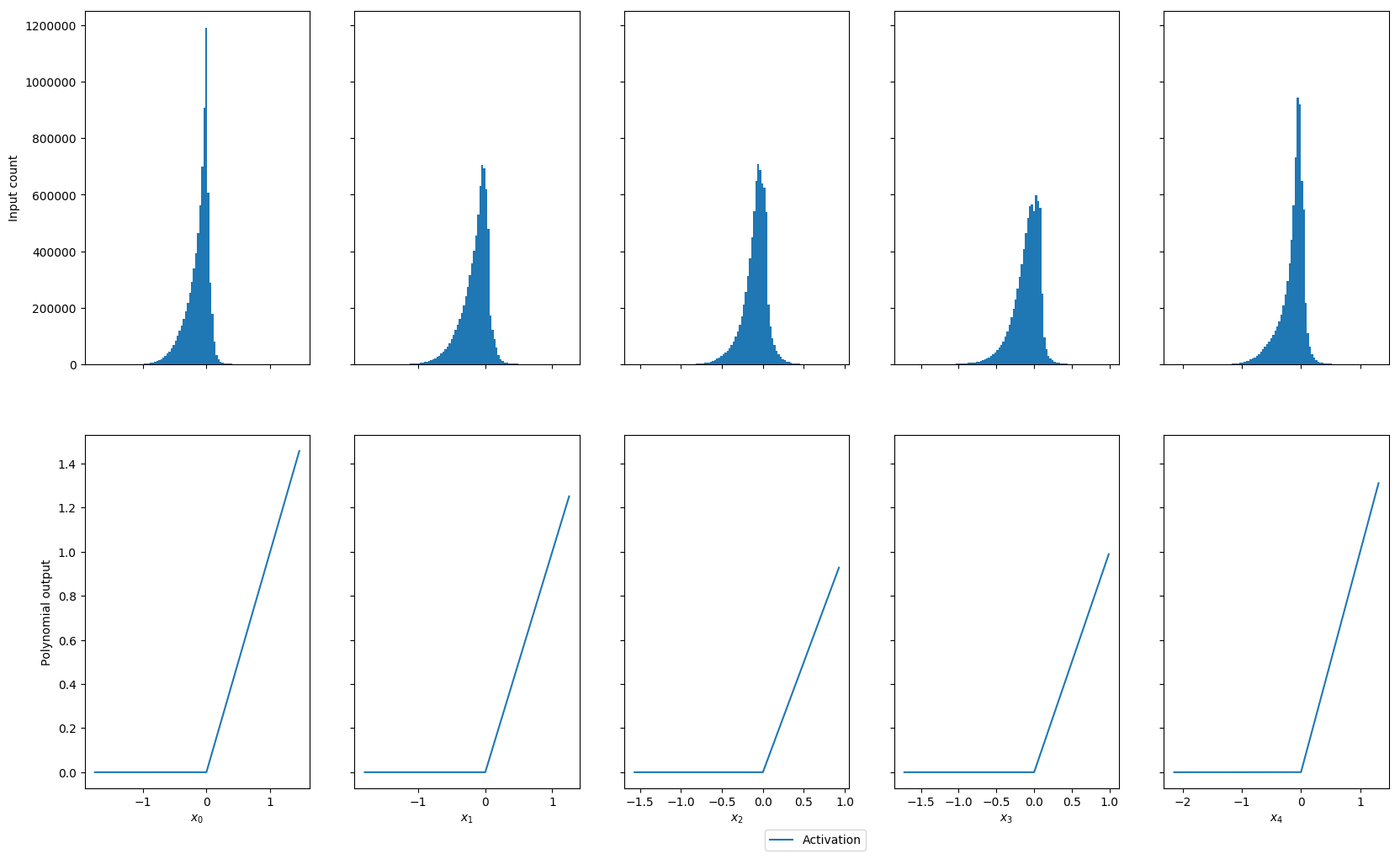}
            \label{relu-hist-deep}
        }
        \caption{
            Sample plots of ReLU and histograms of their inputs for the first 5 hidden units at the 6th (\ref{relu-hist-shallow}) and 13th (\ref{relu-hist-deep}) convolutional layers of the deep residual network trained on CIFAR-10 classification for 100 epochs.
        }
    
    \label{qualitative-relu}
\end{figure}

\end{document}